\documentclass[runningheads]{waica}

\usepackage{graphicx}
\usepackage{booktabs}
\usepackage[utf8]{inputenc}
\usepackage[T1]{fontenc}
\usepackage{hyperref}
\usepackage{url}
\usepackage{booktabs}
\usepackage{amsfonts}
\usepackage{nicefrac}
\usepackage{microtype}
\usepackage{xcolor}
\usepackage{bbding}
\usepackage{enumitem}
\usepackage{comment}
\usepackage{adjustbox}
\usepackage{makecell}
\usepackage{tabularray}
\UseTblrLibrary{booktabs}
\usepackage{xcolor}
\usepackage{colortbl}
\usepackage{multirow}
\usepackage{stfloats}
\usepackage{float}
\usepackage[most]{tcolorbox}
\newtcolorbox{AIbox}[1][]{
  colback=blue!10, 
  colframe=black!100, 
  coltitle=white,   
  fonttitle=\bfseries,  
  title=#1,       
  enhanced,         
  boxrule=0.8pt,     
  arc=1mm,       
  left=6pt, right=6pt, top=6pt, bottom=6pt, 
  breakable   
}

\begin{document}

\title{RoboFAC: A Comprehensive Framework for Robotic Failure Analysis and Correction}

\titlerunning{RoboFAC}

\author{
Zewei Ye\inst{1,*} \and
Weifeng Lu\inst{1,*} \and
Minghao Ye\inst{1,*} \and \\
Tao Lin\inst{1} \and
Ruihan Tao\inst{1} \and
Shuo Yang\inst{2} \and
Junchi Yan\inst{1} \and
Bo Zhao\inst{1}\textsuperscript{\(\dagger\)}
}

\authorrunning{Z. Ye et al.}

\institute{
School of AI, Shanghai Jiao Tong University
\and
Harbin Institute of Technology, Shenzhen
\\
\textsuperscript{*}Equal contribution.
\qquad
\textsuperscript{\(\dagger\)}Corresponding author: \email{bo.zhao@sjtu.edu.cn}
}
\maketitle

\begin{abstract}
Vision-Language-Action (VLA) models have recently advanced robotic manipulation by translating natural-language instructions and visual observations into control actions. 
However, existing VLAs are primarily trained on successful expert demonstrations and lack structured supervision for failure diagnosis and recovery, limiting robustness in open-world scenarios.
To address this limitation, we propose the Robotic Failure Analysis and Correction (\textit{RoboFAC}) framework.
We construct a large-scale failure-centric dataset comprising 9,440 erroneous manipulation trajectories and 78,623 QA pairs across 53 scenes in both simulation and real-world environments, with systematically categorized failure types.
Leveraging this dataset, we develop a lightweight multimodal model specialized for task understanding, failure analysis, and failure correction, enabling efficient local deployment while remaining competitive with large proprietary models. 
Experimental results demonstrate that RoboFAC achieves a 34.1\% higher failure analysis accuracy compared to GPT-4o.
Furthermore, we integrated RoboFAC as an external supervisor in a real-world VLA control pipeline, yielding a 29.1\% relative improvement across four tasks while significantly reducing latency relative to GPT-4o.
These results demonstrate that RoboFAC enables systematic failure diagnosis and recovery, significantly enhancing VLA recovery capabilities.
Our project and dataset are publicly available at \url{https://github.com/MINT-SJTU/RoboFAC}.
\end{abstract}
\section{Introduction}

Vision-Language-Action (VLA) models have demonstrated impressive generalization in robotic manipulation~\cite{kim2024openvlaopensourcevisionlanguageactionmodel,black2024pi0visionlanguageactionflowmodel,cai2024spatialbot,dai2026conlacontrastivelatentaction,lin2025evo1lightweightvisionlanguageactionmodel,intelligence2025pi,nvidia2025gr00tn1openfoundation}. By grounding language instructions into actions via visual feedback, these models handle various tasks. However, in complex, long-horizon scenarios, physical disturbances and execution errors are almost inevitable. Because most VLA models are trained predominantly on expert demonstrations, they lack critical exposure to failure trajectories. Consequently, when an anomaly occurs, these models lack the intrinsic capability to detect the error, diagnose its root cause, and re-plan effectively, often resulting in brittle behaviors and cascading failures~\cite{hu2025racrobotlearninglonghorizon,lin2026onetwovlaunifiedvisionlanguageactionmodel,xia2025phoenixmotionbasedselfreflectionframework}. 

\begin{table*}[htbp]
  \centering
  \caption{\textbf{Comparison of robot manipulation failure QA datasets. }We evaluate existing datasets based on failure taxonomies, video availability, high/low-level correction questions, task horizon/dynamics, and multi-dimensional analysis.}
  \label{tab:comparison}
  \adjustbox{width=\textwidth}{
  \begin{tabular}{lcccccccc}
    \toprule
    \textbf{Datasets} & \makecell{\textbf{Failure} \\ \textbf{Taxonomies}} & \makecell{\textbf{Videos}} & \makecell{\textbf{High-level} \\ \textbf{correction}} & \makecell{\textbf{Low-level} \\ \textbf{correction}} & \makecell{\textbf{Long-horizon} \\ \textbf{Tasks}} & \makecell{\textbf{Dynamic} \\ \textbf{Tasks}} & \makecell{\textbf{Multi-dimensional} \\ \textbf{Analysis}} \\
    \midrule
    RoboFail~\cite{liu2023reflect} 
    & 8 & \textcolor{green!50!black}{\Checkmark} & \textcolor{green!50!black}{\Checkmark} & \textcolor{red}{\XSolidBrush} & \textcolor{green!50!black}{\Checkmark} & \textcolor{red}{\XSolidBrush} & \textcolor{red}{\XSolidBrush} \\
    AHA dataset~\cite{duan2024ahavisionlanguagemodeldetectingreasoning} 
    & 7 & \textcolor{red}{\XSolidBrush} & \textcolor{red}{\XSolidBrush} & \textcolor{red}{\XSolidBrush} & \textcolor{red}{\XSolidBrush} & \textcolor{red}{\XSolidBrush} & \textcolor{red}{\XSolidBrush} \\
    RACER dataset~\cite{dai2024racerrichlanguageguidedfailure} 
    & 2 & \textcolor{red}{\XSolidBrush} & \textcolor{red}{\XSolidBrush} & \textcolor{green!50!black}{\Checkmark} & \textcolor{red}{\XSolidBrush} & \textcolor{red}{\XSolidBrush} & \textcolor{red}{\XSolidBrush} \\
    Guardian dataset~\cite{pacaud2025guardiandetectingroboticplanning} 
    & 11 & \textcolor{red}{\XSolidBrush} & \textcolor{red}{\XSolidBrush} & \textcolor{red}{\XSolidBrush} & \textcolor{red}{\XSolidBrush} & \textcolor{red}{\XSolidBrush} & \textcolor{red}{\XSolidBrush} \\
    \midrule
    \textbf{RoboFAC dataset (Ours)}
    & 6 & \textcolor{green!50!black}{\Checkmark} & \textcolor{green!50!black}{\Checkmark} & \textcolor{green!50!black}{\Checkmark} & \textcolor{green!50!black}{\Checkmark} & \textcolor{green!50!black}{\Checkmark} & \textcolor{green!50!black}{\Checkmark} \\
    \bottomrule
  \end{tabular}
  }
\end{table*}

To address this vulnerability, decoupling execution from failure reasoning offers a modular paradigm to enhance robustness without retraining the base policy. While general-purpose Multimodal Large Language Models (MLLMs) show strong reasoning capabilities, directly deploying them as external critics is challenging: First, they often falter in specialized robotic domains because they are not trained on fine-grained, physics-grounded manipulation failures~\cite{li2025selfcorrectingvisionlanguageactionmodelfast,liu2023reflect,xiong2024aicmllmautonomousinteractive,chen2024automatingrobotfailurerecovery}. Second, their massive parameter scale and high API latency severely hinder real-time, on-device deployment for reactive control~\cite{mirchandani2023largelanguagemodelsgeneral,sinha2024realtimeanomalydetectionreactive}. Although existing specialized datasets for robot failure analysis~\cite{duan2024ahavisionlanguagemodeldetectingreasoning,dai2024racerrichlanguageguidedfailure,pacaud2025guardiandetectingroboticplanning} attempt to alleviate some domain-shift issues, they remain constrained by simplistic task setups, coarse-grained diagnostics, and a lack of multi-level corrective strategies (Table~\ref{tab:comparison}).

To overcome these limitations, we introduce \textbf{RoboFAC}, a unified framework for robotic failure analysis and correction. As illustrated in Figure~\ref{fig:overview}, the foundation of our approach is the \textbf{RoboFAC Dataset}, a large-scale collection of manipulation failures spanning both simulated and real-world environments. To ensure broad coverage of robotic failure modes, we define a taxonomy of six fundamental error types, spanning multiple levels from \textit{task planning} and \textit{motion planning} to \textit{execution control}. This structured decomposition enables the dataset to capture diverse and representative failure patterns, providing a strong foundation for training MLLMs to analyze and reason about errors. 

Leveraging this structured dataset, we develop the \textbf{RoboFAC model}, a lightweight MLLM tailored for failure video reasoning. By explicitly training on our hierarchical data, this open-source model seamlessly integrates task understanding, failure diagnosis, and corrective reasoning, overcoming the limitations of base models lacking failure-specific priors.

Extensive empirical evaluations demonstrate the efficacy of our framework. Despite its compact 7B parameter scale, RoboFAC model substantially outperforms its pre-trained base model and achieves performance comparable to---and in some scenarios surpassing---proprietary general-purpose models like GPT-4o in failure reasoning. Crucially, when integrated as an external critic in real-world VLA pipelines, RoboFAC not only improves task success rates through effective failure recovery, but also reduces inference latency by approximately $3\times$ compared to GPT-4o-based systems, enabling highly responsive, on-device deployment.

Our primary contributions are summarized as follows:
\begin{enumerate}%[itemsep=0pt, topsep=0pt, parsep=0pt, partopsep=0pt]
    \item We introduce the \textbf{RoboFAC Dataset}, a large-scale, hierarchically annotated video QA dataset comprising 78K pairs across eight distinct question types. It systematically categorizes failures across the control hierarchy and incorporates diverse visual variations to support robust, comprehensive failure understanding and correction.
    \item We develop the \textbf{RoboFAC model}, a lightweight, deployable MLLM specialized for robotic failure reasoning. It serves as a unified critic for real-time failure detection and recovery in VLA systems.
    \item We demonstrate that integrating RoboFAC into real-world manipulation pipelines yields higher recovery success rates and a $3\times$ reduction in inference latency compared to state-of-the-art proprietary models like GPT-4o.
\end{enumerate}

\section{Related Work}

\subsection{Robot Manipulation with VLA}

Vision-Language-Action (VLA) models have emerged as a powerful paradigm in Embodied AI, connecting multimodal perception with robotic action generation~\cite{black2024pi0visionlanguageactionflowmodel,cheang2024gr2generativevideolanguageactionmodel,kim2024openvlaopensourcevisionlanguageactionmodel,zhang2025image}. 
By representing robot actions as text tokens, OpenVLA~\cite{kim2024openvlaopensourcevisionlanguageactionmodel} unifies the modalities of vision, language, and action, enabling the model to leverage pre-trained vision-language models for robotic control.  $\pi_0$~\cite{black2024pi0visionlanguageactionflowmodel} further advances this direction by using flow-matching diffusion to decode hidden representations into continuous actions. Other models, such as GR-2~\cite{cheang2024gr2generativevideolanguageactionmodel}, adopt a two-stage training paradigm: pre-training on large-scale internet videos to learn general world dynamics, followed by fine-tuning on robot trajectories for action prediction and video generation. This approach enables GR-2 to generalize effectively across diverse manipulation tasks and environments.
Despite these advances, existing VLAs often exhibit limitations in multi-step tasks requiring temporal reasoning. For example, long-horizon instructions may be misinterpreted due to temporal delays, leading to incorrect grasps or skipped subgoals. In dynamic environments, action trajectories may deviate from intended targets due to accumulated prediction errors.
To address these limitations, we train an auxiliary model to assist VLAs by detecting, analyzing, and correcting failures in real time, thereby enhancing their robustness in complex manipulation tasks.

\subsection{Robot Failure Detection and Analysis}

While Vision-Language-Action (VLA) models have shown remarkable progress in end-to-end robotic control, they often struggle to detect and recover from failures autonomously in unstructured environments. To mitigate these shortcomings, recent work has explored the use of Multimodal Large Language Models (MLLMs) as auxiliary agents for error detection and reasoning. 
MLLMs excel at understanding visual content and producing structured explanations, making them well-suited for post-hoc or real-time failure analysis in manipulation tasks~\cite{cai2025oscnet,luo2025roboreflectroboticreflectivereasoning,duan2024ahavisionlanguagemodeldetectingreasoning,shi2024yellrobotimprovingonthefly,zhou2025codeasmonitorconstraintawarevisualprogramming,dai2024racerrichlanguageguidedfailure,li2025selfcorrectingvisionlanguageactionmodelfast,cai2024spatialbot,pacaud2025guardiandetectingroboticplanning}.
However, many general-purpose MLLMs~\cite{yang2023dawnlmmspreliminaryexplorations,geminiteam2024geminifamilyhighlycapable} are not specifically fine-tuned on robot manipulation data and thus often struggle to accurately analyze operational errors in robotic systems. 
To address this limitation, Luo et al.~\cite{luo2025roboreflectroboticreflectivereasoning} adopt Chain-of-Thought (CoT) prompting strategies to guide the reasoning process within powerful vision-language models, incorporating iterative model calls to ensure consistency in failure diagnosis. 
Shi et al.~\cite{shi2024yellrobotimprovingonthefly} introduce human-in-the-loop feedback mechanisms that collect corrective data during robot execution and use it for model fine-tuning. 
Dai et al.~\cite{dai2024racerrichlanguageguidedfailure} and Duan et al.~\cite{duan2024ahavisionlanguagemodeldetectingreasoning} construct image-text datasets centered on failure cases in manipulation, enabling supervised training of MLLMs for error detection.
In contrast, we propose a video-based dataset for robotic failure analysis and correction, encompassing tasks from short to long horizons. Building on our dataset, we fine-tune a dedicated MLLM that achieves accurate and fine-grained failure understanding and recovery. This enables more robust and transparent deployment of vision-language models in diverse and challenging robotic manipulation scenarios.

\section{The RoboFAC Dataset}
\label{Section:dataset}
In this section, we introduce the RoboFAC dataset, which is a large-scale and diverse dataset for question-answering on robot failure videos. We begin with an overview of the RoboFAC dataset, followed by a detailed definition of the failure taxonomies included in the dataset. Finally, we present how we constructed the RoboFAC dataset.

\subsection{Overview of the RoboFAC Dataset}

\begin{figure*}[t]
    \centering
    \includegraphics[width=\textwidth]{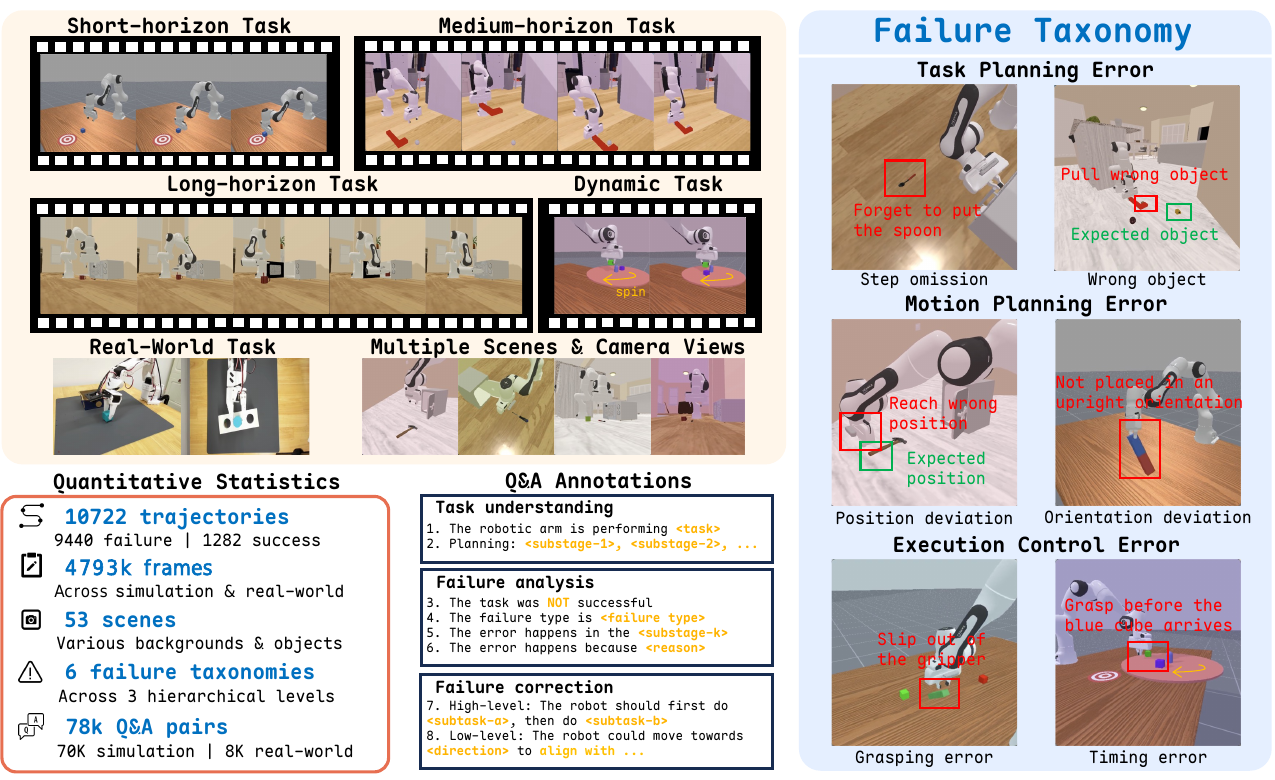}
    \caption{
    Overview of RoboFAC dataset. \textbf{Left:} The RoboFAC dataset features both task diversity and visual diversity, encompassing tasks of varying complexity, real-world tasks, and various backgrounds and camera viewpoints. We provide detailed video question-answer annotations for eight distinct question types. \textbf{Right:} A detailed visual illustration of the six failure taxonomies.
    }
    \label{fig:overview}
\end{figure*}

\begin{figure*}[htbp]
    \centering
    \includegraphics[width=\linewidth]{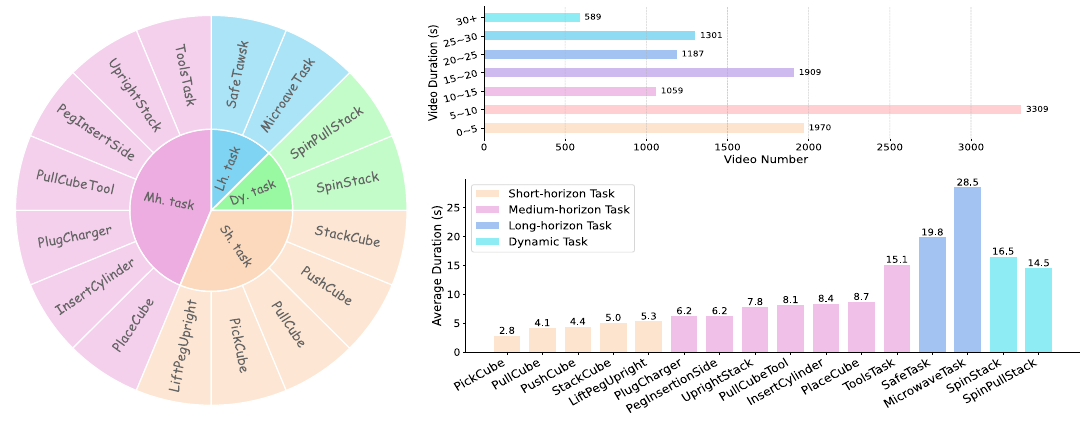}
    \caption{Statistics of the RoboFAC Dataset. \textbf{Left:} Categories of robotic tasks in the RoboFAC dataset. (Lh. Task: Long-horizon task, Mh. Task: Medium-horizon task, Sh. Task: Short-horizon Task, Dy. Task: Dynamic Task) \textbf{Top Right:} Distribution of video counts by duration interval. \textbf{Bottom Right:} Average duration of each task.}
    \label{fig:dataset statistics}
\end{figure*}

The RoboFAC dataset covers robotic tasks of varying complexity, ranging from short-horizon to long-horizon scenarios, and includes dynamic environments (Figure~\ref{fig:dataset statistics}, left). It comprises 14 simulated tasks and 6 real-world tasks, two of which are not available in simulation. The dataset captures six types of failures across three hierarchical levels (Section~\ref{section:Taxonomy of Failures}).

To reflect diverse real-world deployment conditions, we introduce variations in backgrounds and camera viewpoints, significantly enhancing visual diversity. This design improves visual generalization and enables more robust evaluation.

In total, RoboFAC contains 8,960 simulated and 480 real-world failure trajectories. To mitigate overfitting to failure patterns, we additionally include 1,160 successful simulated trajectories and 122 real-world ones. After annotation, the dataset yields 78K video QA samples, including 70K from simulation and 8K from real-world data.

\subsection{Taxonomy of Failures}
\label{section:Taxonomy of Failures}

We propose a three-level taxonomy of failures in robotic manipulation, inspired by prior analyses ~\cite{liu2023reflect,duan2024ahavisionlanguagemodeldetectingreasoning} and aligned with a hierarchical task structure (Figure~\ref{fig:overview}, Right): \emph{Task Planning}, \emph{Motion Planning}, and \emph{Execution Control}, inspired by classic robotics literature~\cite{siciliano2009robotics} . Each level abstracts a distinct source of error, enabling targeted diagnosis and remediation. In addition, these failures, being atomic and task-independent, can be consistently observed during robot manipulation and occur frequently in our experiments.

Assume a task $T$ is composed of substages $\{S_i\}_{i=1}^{N}$, where each substage involves the execution time $t$, the end-effector’s position $p \in \mathbb{R}^3$, orientation denoted by a unit quaternion $q$, gripper closure level $G\in[0,1]$, and the manipulated object $b\in\mathcal B$, where $\mathcal B=\{b_1, ..., b_M\}$ is the set of all the objects in the environment.
Ideally, the actual execution parameters $(\tilde{p}_i, \tilde{q}_i, \tilde{G}_i, \tilde{b}_i, \tilde{t}_i)$ at substage $S_i$ should match the correct parameters $(p_i, q_i, G_i, b_i, t_i)$, ensuring successful task completion.
However, errors occur when any of these parameters deviate from nominal values, causing the task to fail. We define the failure taxonomy as follows:

\subsubsection{Task Planning Error.}
Errors rooted in incorrect task \textit{decomposition} or failed language grounding in VLA models.

\textit{Step Omission:} A required substage $S_i$ is skipped, resulting in an incomplete plan: $(S_1, ..., S_{k-1}, S_{k+1}, ..., S_N)$. 

\textit{Wrong Object:} Fail to select the correct object to manipulate as specified by the language instruction: $\tilde{b_i} \in \mathcal B \setminus {b_i}$.

\subsubsection{Motion Planning Error.}
Failures arising from limited spatial reasoning or inaccurate mapping from instructions to poses. This causes the current subtask to fail.

\textit{Position Deviation:} The end-effector fails to reach the correct position. $\tilde{p}_i = p_i + \delta p_i$, with $\delta p_i \in \mathbb{R}^3$.

\textit{Orientation Deviation:} The end-effector fails to reach the correct orientation. $\tilde{q}_i = \delta q_i\otimes q_i$, where $\delta q_i$ is a unit quaternion and $\otimes$ represents quaternion multiplication.

\subsubsection{Execution Control Error.}
Execution control failures caused by physical imprecision, latency, or dynamic misalignment during actuation and environment interaction.

\textit{Grasping Error:} The gripper does not close properly or the closure level is insufficient: $\tilde{G_i} < G_i$. This results in failure to grasp the target object or causes the object to slip from the gripper.

\textit{Timing Error:} Executing the subtask at an incorrect timing. $\tilde{t}_i = t_i\pm\delta t$, where $\delta_t$ introduces temporal offsets.

% ---
\subsection{Data Construction Pipeline}  

Our data construction pipeline consists of \textit{data collection} and \textit{data annotation}, followed by a three-stage \textit{quality control process} with human verification.

\subsubsection{Data Collection.} 
\label{Section3.2.1 Data collection}
\begin{figure*}[htbp]
    \centering
    \includegraphics[width=\linewidth]
    {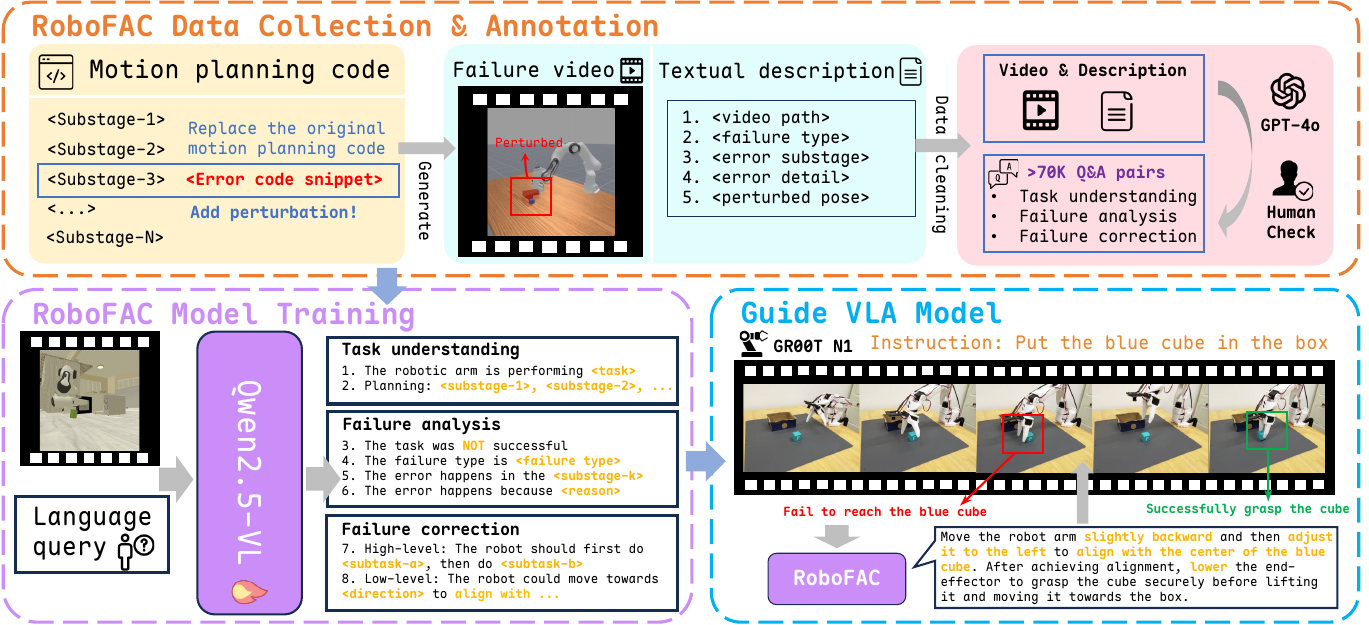}
    \caption{Overview of our RoboFAC framework. \textbf{Top:} The pipeline of constructing the RoboFAC dataset. \textbf{Bottom-left:} We build our RoboFAC model by fine-tuning Qwen2.5-VL model. The RoboFAC model can perform Task Understanding, Failure analysis and Failure correction. \textbf{Bottom-right:} We deploy RoboFAC model on real-world VLA control tasks, and it effectively helps the VLA recover from failure.}
    \label{fig:pipeline}
\end{figure*}

\textbf{Simulation Data.} 
Our dataset construction pipeline in the simulation environment is illustrated at the top of Figure~\ref{fig:pipeline}. We collect the simuluation data for 14 robotic tasks in the ManiSkill environment~\cite{tao2024maniskill3gpuparallelizedrobotics}, 
augmented with objects from the YCB Object Dataset~\cite{Calli_2015} to increase object diversity and scenes from ReplicaCAD~\cite{szot2022habitat20traininghome} and AI2-THOR~\cite{kolve2022ai2thorinteractive3denvironment} to enrich environmental diversity.
For each task, we define an expert policy via target end-effector poses at each substage, with feasible trajectories generated through motion planning. Failure data is constructed by replacing the expert policy with a code snippet that introduces erroneous trajectories at selected substages, leading to task failure.

Each failure trajectory is recorded as a video with a textual description specifying the failure substage, taxonomy, and detailed explanation. For pose perturbation failures, the perturbed pose is also recorded. These descriptions are later used for annotation.

\textbf{Real-World Data.} 
We collect real-world data for 6 tasks, including two not present in simulation, using teleoperation with the SO-100 robotic arm. Each video is paired with a corresponding textual description.

\subsubsection{Data Annotation.}
\label{sec:data_annotation}
We annotate the raw data to construct video-based QA samples corresponding to eight question types, which are described in detail in Section~\ref{section: model}. These eight question types comprehensively evaluate a model's ability in \textbf{Task Understanding}, \textbf{Failure Analysis}, and \textbf{Failure Correction} based on robot manipulation videos. For each type, we provide five question templates.

Reference answers are derived from textual descriptions. For five question types--\textit{task identification}, \textit{task planning}, \textit{failure detection}, \textit{failure identification}, and \textit{failure locating}--the reference answers can be directly extracted from the corresponding textual description, as they have well-defined ground truths. For the remaining three types-\textit{failure explanation}, \textit{high-level correction}, and \textit{low-level correction}-we utilize both the video and its corresponding textual description as inputs to GPT-4o to generate the reference answers.

\subsubsection{Quality Control Process.}

We adopt a three-stage quality control pipeline: simulation validation, LLM-based annotation verification, and human consistency evaluation.

\textbf{Simulation Validation.}
During motion planning, we enforce physical validity constraints to eliminate spurious failures. Specifically, we perform (1) unexpected environment collision detection, and discard trajectories where the environmental state changes unexpectedly; and (2) trajectory discontinuity detection by examining joint-level temporal differences to remove non-smooth trajectories. Only physically valid and temporally consistent trajectories are retained.

\textbf{LLM-Based Annotation Validation.}
Annotations are generated using a fixed prompt template and required to follow a predefined JSON schema. By applying automatic parsing and validation, invalid outputs are filtered before human review.

\textbf{Human Verification and Consistency.}
All samples in the dataset are manually annotated using a four-level ordinal quality scale and subsequently screened to ensure annotation correctness and data quality. To assess annotation consistency, 10\% of the dataset is independently rated by two randomly selected annotators using the same scale. Inter-annotator agreement is measured using Krippendorff’s $\alpha$~\cite{Krippendorff1980ContentAA} with an ordinal metric, yielding $\alpha = 0.86$, indicating high reliability.
\section{The RoboFAC Model}

\label{section: model}
This section introduces our \textbf{RoboFAC model}, which demonstrates strong capabilities in \textbf{Task Understanding}, \textbf{Failure Analysis}, and \textbf{Failure Correction}. As illustrated in the bottom-left corner of Figure~\ref{fig:pipeline}, given a robot manipulation video, the model is able to comprehensively interpret the video in natural language in a video-question-answering (VideoQA) manner. 

\noindent \textbf{Task Understanding.} This capability is to understand the robotic task through the video, encompassing both \textit{task identificaiton} and \textit{task planning}. Specifically, given a robot manipulation video $\mathcal V$, the model identifies what the robot is doing through the video as task $T$, and decomposes the task into a sequence of substages $(S_1, S_2, \dots, S_N)$ by analyzing how the robot performs the task in the video.

\noindent \textbf{Failure Analysis.} Our model is able to conduct comprehensive analyses of failures in robot manipulation videos, including:
\begin{itemize}[itemsep=0pt, topsep=0pt, parsep=0pt, partopsep=0pt]
    \item \textit{Failure detection:} Determine whether the robotic task in the video was successfully completed.
    \item \textit{Failure identification:} If the robotic task fails, determine what is the type of the failure.
    \item \textit{Failure locating:} If the robotic task fails, determine in which step the error happens.
    \item \textit{Failure explanation:} If the robotic task fails, provide detailed explanation for the failure happened in the video.
\end{itemize}

\noindent \textbf{Failure Correction.} Our RoboFAC model provides detailed correction suggestions for errors observed in the video, helping the VLA model recover from failures. These suggestions come in two forms: \textit{high-level corrections} specify the sequence of sub-tasks for recovery, useful for task-planning errors such as missing or misordered sub-tasks; \textit{low-level corrections} offer fine-grained guidance on end-effector movement direction, better suited for execution-level errors such as inaccurate positioning or unsuitable trajectories. We extensively validate this capability in real-world scenarios, with detailed settings and results in Section~\ref{sec:real_exp}.

\noindent \textbf{Model Architecture.} We build our model on Qwen2.5-VL~\cite{bai2025qwen25vltechnicalreport}, one of the most advanced open-source multi-modal models to date, consisting of an LLM backbone, a vision encoder, and an MLP-based vision-language merger. Qwen2.5-VL model supports single-image, multi-image, and video inputs at varying resolutions, achieving strong performance in visual question answering tasks. Our further training details are provided in Section~\ref{sec:robofac_setup}.
\section{Experiments}

In this section, we comprehensively evaluate our model's capacity. We compare our model against both proprietary and open-source models on failure analysis task across multiple performance dimensions. Additionally, we deploy our model as a critic to supervise a real-world robotic arm during task execution, assessing whether it can effectively guide the VLA model and thus enhance the success rate of robotic tasks in real-world scenarios.

\subsection{Experiment Setup on Failure Analysis Task}
\label{sec:robofac_setup}
\noindent \textbf{Training Set \& Failure Analysis Task.}
We construct the training and evaluation datasets from our collected RoboFAC data. Specifically, we randomly sample 60K QA pairs from the simulated RoboFAC dataset for training. The remaining data is used for evaluation, including 10K simulated QA pairs and 8K real-world QA pairs. Notably, the simulated test split contains over 1,000 robotic videos that are entirely unseen during training. Moreover, the model is never trained on real-world data, and the real-world test split includes two unseen tasks (InsertCylinder and PlaceCube). This setup enables a rigorous evaluation of sim-to-real transfer and generalization ability.

\noindent \textbf{Training Details.}
Pretrained VLMs are not tailored for robotic manipulation, leading to a domain gap in robotic visual question answering. To mitigate this, we freeze the visual encoder to retain general visual representations, while fully fine-tuning the merger and LLM backbone. The merger learns robot-specific alignment between visual features and language embeddings, while the LLM adapts to incorporate embodied visual signals into reasoning.

Concretely, we fine-tune Qwen2.5-VL-3B and Qwen2.5-VL-7B on the RoboFAC training set for one epoch with a learning rate of $1\times10^{-5}$. We unfreeze both the LLM backbone and merger modules. We adopt DeepSpeed ZeRO-3 offloading~\cite{rajbhandari2020zero} for memory efficiency, using a per-GPU batch size of 1 and gradient accumulation steps of 2 and 4 for the 3B and 7B models, respectively. Training is conducted on 4 NVIDIA RTX 4090 GPUs, requiring approximately 10 hours (3B) and 24 hours (7B).

\noindent \textbf{Evaluation Metrics.}
We adopt different metrics for different task types. For \textit{failure detection}, \textit{failure identification}, and \textit{failure localization}, where answers are deterministic, we use a multiple-choice setting and report accuracy. For other tasks with open-ended responses, we employ an external LLM to evaluate answers along three dimensions: \textbf{correctness}, \textbf{relevance}, and \textbf{completeness}. The final score is the average of these three dimensions, normalized to a 100-point scale.

\subsection{Results on Failure Analysis Task}
\label{sec:result}

\begin{figure*}[htbp]
    \centering
    \includegraphics[width=\linewidth]{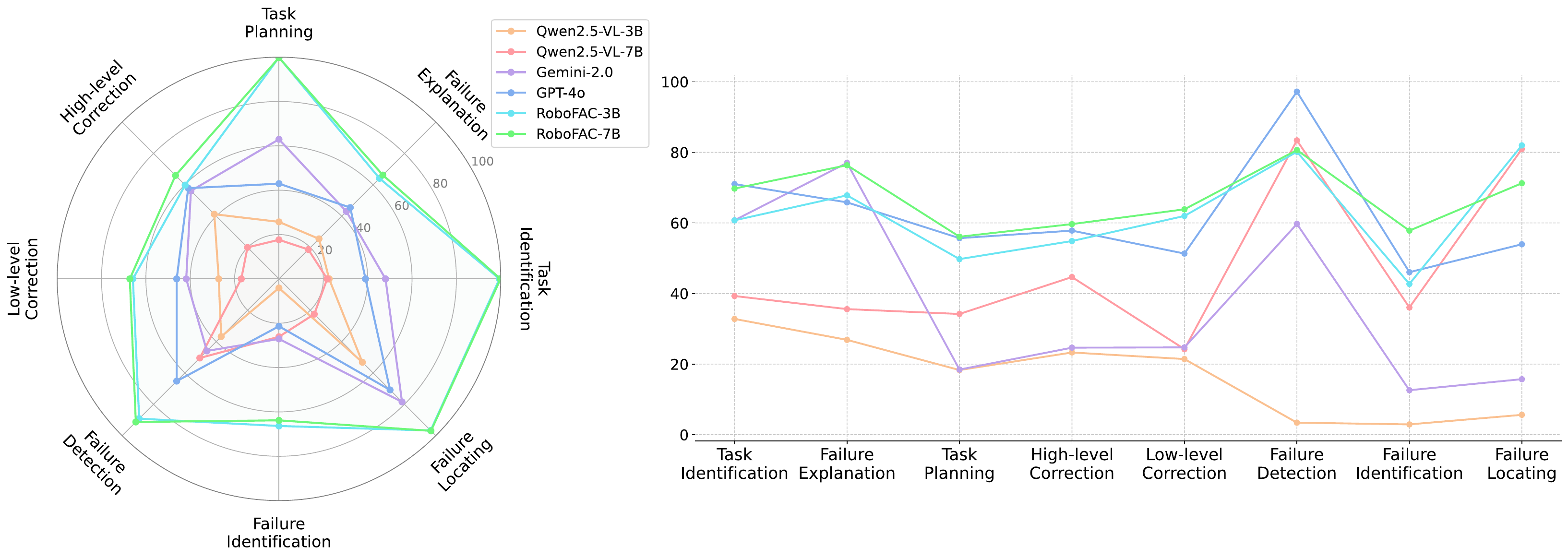}
    \caption{Scores for different dimensions on RoboFAC Benchmark \textbf{Left:} Performance on different question dimensions for simulation dataset. \textbf{Right:} Performance on different question dimensions for real world dataset.}
    \label{fig:result}
\end{figure*}

We comprehensively evaluate our proposed RoboFAC models (RoboFAC-3B and RoboFAC-7B) against several strong multimodal baselines, including open-source models Qwen2.5-VL-3B and Qwen2.5-VL-7B, and proprietary models Gemini-2.0 and GPT-4o. The evaluation spans diverse manipulation tasks and cognitive abilities essential for robotic reasoning, with metrics defined in Section \ref{sec:robofac_setup}. The results are summarized in Figure~\ref{fig:result} and Table~\ref{tab:result}.

\begin{table*}[htbp]
\centering
\caption{Performance of various multimodal models on the failure analysis tasks. The scores represent success rates (\%).}
\label{tab:result}
    \adjustbox{max width=\textwidth}{
        \begin{tabular}{lcccccc}
        \toprule
        \multicolumn{1}{c}{\textbf{Model}} & \makecell{\textbf{Short-horizon} \\ \textbf{Task}} & \makecell{\textbf{Medium-horizon} \\ \textbf{Task}} & \makecell{\textbf{Long-horizon} \\ \textbf{Task}} & \makecell{\textbf{Dynamic} \\ \textbf{Task}} & \makecell{\textbf{Real-world} \\ \textbf{Task}} & \textbf{Average}        \\
        \midrule
        Qwen-2.5-VL-3B            & 40.99              & 27.82               & 25.18             & 28.94          & 17.36           & 27.82          \\
        Qwen-2.5-VL-7B            & 14.26              & 11.73               & 38.84             & 18.00          & 50.96           & 27.47          \\
        Gemini-2.0                & 63.32              & 53.23               & 45.67             & 48.91          & 41.72           & 51.11          \\
        GPT-4o                    & 61.50              & 53.81               & 42.46             & 45.82          & 65.89           & 57.42          \\
        RoboFAC-3B                & 81.66              & 84.67               & 79.32             & 83.02 & 63.29           & 76.80          \\
        RoboFAC-7B                & \textbf{82.74}     & \textbf{84.92}      & \textbf{81.78}    & \textbf{83.28} & \textbf{68.94}  & \textbf{79.10} \\
        \bottomrule
        \end{tabular}
    }
\end{table*}

\noindent \textbf{Overall Performance.}
As shown in Table~\ref{tab:result}, RoboFAC-7B consistently outperforms all baselines across 
%all kinds of tasks,
short-, medium-, long-horizon, dynamic, and real-world tasks, 
achieving an average score of \textbf{79.10}—significantly surpassing GPT-4o (\textbf{57.42}) and Gemini-2.0 (\textbf{51.11}). Notably, even the smaller RoboFAC-3B reaches \textbf{76.80}, highlighting the effectiveness of our domain-specific training and architectural design.

\noindent \textbf{Multi-Dimensional Capacity.}
Figure~\ref{fig:result} further breaks down the performance across eight key capacities critical to robotic failure comprehension: task understanding (task identification, task planning, failure correction (high/low level), and failure analysis (detection, identification, locating, explanation). Our RoboFAC model demonstrates a strong ability to handle robotic failures, achieving the highest or near-highest scores in task planning, low-level correction, and all three failure-related abilities. This indicates that our models are capable of nuanced task decomposition and resilient recovery from execution failures, both of which are essential for real-world deployment.

In contrast, large-scale generalist models such as GPT-4o and Gemini-2.0, while competitive in certain aspects (e.g., failure detection), exhibit limited performance in task planning and hierarchical correction. This suggests a gap in their ability to perform complex, multi-step reasoning under physical constraints, which our models are specifically trained to address.

\noindent \textbf{Sim2Real Analysis.}
We further evaluate the sim-to-real transfer of RoboFAC on real-world robotic videos with results shown in the right panel of Figure~\ref{fig:result}). RoboFAC largely retains its simulation-level performance across most task dimensions and consistently outperforms all baselines on the majority of categories. Despite never seeing real-world data during training, both RoboFAC-3B and RoboFAC-7B remain robust under domain shift, showing that the learned failure representations and corrective reasoning generalize beyond simulation.

Notably, generalist models score slightly higher on perception-oriented dimensions such as failure detection and task identification, likely owing to their large-scale visual–language pretraining, which yields stronger priors for semantic recognition under real-world appearance variations. In contrast, RoboFAC's domain-specific fine-tuning prioritizes structured task understanding and hierarchical corrective reasoning over generic visual recognition, leading to superior results in task planning and both high-level and low-level correction. These differences reflect a natural divide between general visual understanding and robotic reasoning, further underscoring RoboFAC's advantage in the decision-making and recovery processes essential for real-world manipulation.

\subsection{Experiment Setup on Real-world Manipulation}
\label{sec:real_exp}
\label{sec:real_world}

Our real-world experiments aim to answer two questions:
\begin{itemize}[itemsep=0pt, topsep=0pt, parsep=0pt, partopsep=0pt]
    \item[(Q1)] Does RoboFAC provide measurable benefits over the original Qwen2.5-VL and the closed-source GPT-4o for robotic error correction?
    \item[(Q2)] What types of corrective instructions are most effective for VLA-based policies, and how does the policy respond to them?
\end{itemize}

\noindent \textbf{Experiment Setup.}
\label{real_setup}
We build a physical evaluation system based on the SO-100 robotic arm. Based on the LeRobot framework~\cite{cadene2026lerobotopensourcelibraryendtoend}, we collect 100 teleoperated demonstrations per task. These demonstrations, together with paired correction-style language instructions and corresponding action traces, are used to fine-tune the GR00T-N1 policy~\cite{nvidia2025gr00tn1openfoundation}.

Deployment follows a server--client architecture: the correction model runs on a server (or GPT-4o API), while the policy runs on a client controlling the robot. Video segments are sent to the server for correction generation, and returned instructions are used to resume execution.
\emph{Latency} is measured as the average time interval between the client sending observations and receiving the generated corrective instruction, including network transmission and model inference time.

\noindent \textbf{Evaluation Protocol.}
Execution starts from an initial prompt. At a predefined step, a third-view video is extracted for correction generation. The resulting instruction is appended to the prompt and execution resumes. Each trial allows up to four correction rounds (five total attempts including the initial attempt). We report success rates after the first and final rounds.

\noindent \textbf{Comparison Settings.}
We evaluate five conditions: 
(1) No Correction, 
(2) GPT-4o, 
(3) Qwen2.5-VL-7B (no fine-tuning), 
(4) RoboFAC-7B (Low-level correction), and 
(5) RoboFAC-7B (High-level correction). 

\subsection{Results on Real-world Manipulation}

\noindent \textbf{Effect of Fine-Tuned RoboFAC (Q1).}
As shown in Table~\ref{tab:realworld}, RoboFAC-7B (Low-level) achieves the highest final success rate (61.25\% after four correction rounds), outperforming GPT-4o (56.25\%) and clearly exceeding both No Correction (47.5\%) and the untuned Qwen2.5-VL-7B (50.0\%).

\begin{table*}[htbp]
    \centering
    \caption{Success rate on real-world manipulation.}
    \label{tab:realworld}
    \adjustbox{width=\textwidth}{
        \begin{tabular}{l c l c c c c c}
            \toprule
            \textbf{Methods} & \textbf{Latency (s)} & \textbf{ } 
            & \textbf{PlaceCube} & \textbf{PushCube} 
            & \textbf{PullCubeTool} & \textbf{StackCube} 
            & \textbf{Average} \\
            \midrule
            
            \multirow{2}{*}{No correction} 
            & \multirow{2}{*}{--} 
            & 1 attempt  & 0.20 & 0.55 & 0.10 & 0.35 & 30.00\% \\
            &  & 5 attempts & 0.40 & 0.70 & 0.20 & 0.60 & 47.50\% \\
            \midrule
            
            \multirow{2}{*}{GPT-4o} 
            & \multirow{2}{*}{24.3 $\pm$ 3.4}
            & 1 attempt  & 0.25 & 0.70 & 0.15 & 0.50 & 40.00\% \\
            &  & 5 attempts & 0.50 & 0.80 & \textbf{0.30} & 0.65 & 56.25\% \\
            \midrule
            
            \multirow{2}{*}{Qwen2.5-VL-7B} 
            & \multirow{2}{*}{6.9 $\pm$ 0.6}
            & 1 attempt  & 0.35 & 0.60 & 0.15 & 0.45 & 38.75\% \\
            &  & 5 attempts & 0.50 & 0.70 & 0.20 & 0.60 & 50.00\% \\
            \midrule
            
            \multirow{2}{*}{RoboFAC-7B (Low)} 
            & \multirow{2}{*}{6.7 $\pm$ 0.5}
            & 1 attempt  & 0.40 & 0.70 & 0.20 & 0.50 & 45.00\% \\
            &  & 5 attempts & \textbf{0.60} & \textbf{0.85} 
            & \textbf{0.30} & \textbf{0.70} & \textbf{61.25\%} \\
            \midrule
            
            \multirow{2}{*}{RoboFAC-7B (High)} 
            & \multirow{2}{*}{7.0 $\pm$ 0.5}
            & 1 attempt  & 0.45 & 0.65 & 0.10 & 0.45 & 41.25\% \\
            &  & 5 attempts & 0.50 & 0.75 & 0.20 & 0.55 & 50.00\% \\
            
            \bottomrule
        \end{tabular}
    }
\end{table*}

These results indicate that (1) fine-tuning on RoboFAC data substantially improves correction effectiveness compared to the unfine-tuned Qwen2.5-VL, validating the contribution of our dataset and training strategy. (2) despite being a much smaller model, RoboFAC surpasses GPT-4o while reducing inference latency by about $\sim$3$\times$, making it more suitable for real-time deployment.

\noindent \textbf{Effect of Instruction Granularity (Q2).}
Low-level (action-level) corrections consistently outperform high-level (task-level) corrections across correction rounds. This indicates that, after task-specific fine-tuning of the VLA policy, most real-world failures arise from execution-level inaccuracies rather than high-level task misunderstanding. Action-level guidance provides more direct and executable feedback for correcting motion.

\noindent \textbf{Observation on Policy Behavior.}
We further observe that the policy exhibits directional sensitivity to corrective language: when instructed to adjust toward a specific region or modify the grasp location, it typically moves in the intended direction. However, it shows limited sensitivity to precise quantitative constraints, indicating limitations in fine-grained control under language guidance.

% \section{Discussion}
\section{Conclusion}
\label{discussion}

We present RoboFAC, a framework for robotic failure analysis and correction consisting of a large-scale, multi-dimensional dataset and a specialized multimodal model. The dataset covers diverse failure scenarios with annotations spanning task understanding, failure diagnosis, and corrective reasoning. 
Built upon this foundation, the RoboFAC model significantly improves failure reasoning over its base model and achieves competitive performance with GPT-4o. When integrated as an external critic in real-world VLA control pipelines, it increases task success rates and enables efficient, low-latency failure recovery, underscoring its practical value for embodied AI applications.

\bibliographystyle{IEEEtran}
\bibliography{references}

\newpage
\appendix
\section*{Appendix}
\section{Task Description}

For each task, we systematically vary the object categories and modify the scene of the environment to promote task generalization. A brief description of the original tasks we defined is shown below.

\begin{table}[htbp]
\centering
\caption{A brief description of the task we defined. The table is divided into four sections according to the type of task, from top to bottom, Dynamic Tasks, Long-horizon Tasks, Medium-horizon Tasks, and Short-horizon Tasks.}
\label{tab:task}
    \adjustbox{max width=\textwidth}{
        \begin{tabular}{lc}
        \toprule
        \textbf{Task}            & \textbf{Description}                                                                                                    \\
        \midrule
        \textbf{SpinStack}       & \makecell{Pick up the cube on the spinning disc\\ and stack it on another cube on the disc.}                                         \\
        \textbf{SpinPullStack}   & \makecell{Pull out the cube on the spinning disc\\ and stack it on another cube on the disc.}                                        \\
        \midrule
        \textbf{MicrowaveTask}   & \makecell{Put the spoon on the table into the cup. Open the door of microwave,\\ put the cup into the microwave and close the door.} \\
        \textbf{SafeTask}        & \makecell{Put the gold bar into the safe, close the door of the safe\\ and rotate the cross knob on the door to lock it.}            \\
        \midrule
        \textbf{ToolsTask}       & \makecell{Choose the correct (L-shaped) tools,\\ grasp it to pull the correct (2-pins) charger and plug it.}                         \\
        \textbf{UprightStask}    & Upright the peg and stack it on the cube.                                                                               \\
        \textbf{PegInsetionSide} & Insert the peg into the hole on the side of the block.                                                                  \\
        \textbf{PullCubeTool}    & Grasp the L-shaped tool and pull the cube by it.                                                                        \\
        \textbf{PlugCharger}     & Grasp the charger and plug it into the receptacle.                                                                      \\
        \textbf{InsertCylinder}  & Upright the cylinder and insert it into the middle hole on the shelf.                                                   \\
        \textbf{PlaceCube}       & Pick up the cube and place it into the box.                                                                            \\
        \midrule
        \textbf{LiftPegUpright}  & Lift the peg and upright it.                                                                                            \\
        \textbf{PickCube}        & Pick the cube to the target position.                                                                                   \\
        \textbf{PullCube}        & Pull the cube to the red and white target.                                                                              \\
        \textbf{PushCube}        & Push the cube to the red and white target.                                                                              \\
        \textbf{StackCube}       & Pick up the cube and stack it on another cube.                                                                          \\
        \bottomrule
        \end{tabular}
    }
\end{table}

\section{Question Template}
\label{question_template}
For each of the eight question types, we design a set of question templates. To enhance the diversity of our questions, we provide five distinct phrasings for each type. During the construction of a specific QA pair, one template is randomly sampled from the corresponding set. The complete list of templates is as follows:
\begin{AIbox}[Question Template]

\textbf{Task identification}\\
1. Please describe the task the robot is performing in the video.\\
2. Based on the video, what task is the robot carrying out? \\
3. Can you identify what task the robot is doing in the provided video?\\ 
4. What is the robot doing in the video? Please describe its task. \\
5. From the video, what task is the robot engaged in?\\

\textbf{Task planning}\\
1. This is a video of a robotic arm performing a task, please break down its execution into a sequence of substages.\\
2. Given the video of a robotic arm doing a task, please plan its actions as a sequence of substages.\\
3. In the video, the robotic arm executes a task. Please break down its execution into a sequence of substages.\\
4. Watch the video of the robotic arm performing a task, please outline the process as a substages sequence.\\
5. Based on the video showing a robotic arm carrying out a task, please generate a sequence of substages for its execution.\\

\textbf{Failure detection}\\
1. This is a video of a robotic arm performing a task, was the task successfully completed?\\
2. Based on the video of the robotic arm executing a task, did it finish the task successfully?
3. In the video, the robotic arm executes a task, can you determine whether it was successful?
4. Please assess if the robotic arm has successfully accomplished the task.
5. In the video, the robotic arm executes a task, was it successful?

\textbf{Failure identification}\\
1. This is a video of a robotic arm performing a task, please identify the type of error that occurred during execution.\\
2. Based on the video of the robotic arm carrying out a task, what type of error took place during the task?\\
3. The robotic arm failed to complete the task, can you specify the type of error that happened?\\
4. Please describe the error type that occurred during the robotic arm's execution of the task.\\
5. From the video of the robotic arm performing a task, what kind of error can be observed during the task?\\

\textbf{Failure locating}\\
1. This is a video of a robotic arm performing a task, please identify the subtask stage where the error occurred.\\
2. This is a video of a robotic arm performing a task, during which subtask did the error happen?\\
3.The robotic arm failed to complete the task, can you locate the specific subtask in which the error occurred?\\
4. Please determine at what subtask stage the error took place in the robotic arm's performance of the task.\\
5. From the video of the robotic arm carrying out a task, identify the phase of the task where the error happened.\\

\textbf{Failure explanation}\\
1. This is a video of a robotic arm performing a task, please explain in detail the reason for the task failure.\\
2. Based on the video, provide a detailed explanation of why the robotic arm failed to complete the task.\\
3. The robotic arm failed to complete the task, can you describe in detail the cause of the failure in the video?\\
4. Please analyze the video and explain thoroughly what led to the failure of the task.\\
5. From the video of the robotic arm executing a task, give a detailed explanation of the reason behind the task failure.\\

\textbf{High-level correction}\\
1. This is a video of a robotic arm performing a task, an error occurred during execution. Please provide high-level corrective instructions to help the robot recover and complete the task successfully.\\
2. Based on the video showing an error during the robotic arm 's execution of a task, give detailed high-level guidance for correcting the error and enabling task completion.\\
3. In this video, an error happened while the robotic arm was performing the task, please suggest high-level recovery steps so the robot can continue and complete the task.\\
4. The robotic arm failed to complete the task, please analyze the error in the robotic arm's task from the video and propose high-level correction actions that would allow successful task completion.\\
5. From the video of the robotic arm failing during the task, provide high-level corrective commands to guide it to recover and finish the task.\\

\textbf{Low-level correction}\\
1. This is a video of a robotic arm performing a task, an error occurred during execution. Please provide low-level corrective commands to help the robot recover and complete the task successfully.\\
2. Based on the video, an error happened while the robot was executing a task, give detailed low-level instructions to correct the issue and allow the task to be finished.\\
3. According to the video of the robotic arm executing a task, please suggest specific low-level recovery actions to enable successful task completion.\\
4. From the video showing an error in the robotic arm's task, provide precise low-level commands for error correction and recovery.\\
5. In the video, an error occurred during the robot's performance of the task, please give low-level control instructions to help it recover and complete the task.
\end{AIbox}

\section{LLM Data Annotation Details}
\label{annotation_prompt}
For the \textit{failure explanation}, \textit{high-level correction}, and \textit{low-level correction} questions, we employed GPT-4o to annotate the data. Specifically, we constructed prompts using the description files obtained during video collection. We use the prompt paired with the corresponding videos to request GPT-4o.
The constructed prompt is as follows:

\begin{AIbox}[Prompt for data annotation]
This is a video of a robot arm performing a task, and the task is failed.\\

Here is the basic information of the video:\\
- Task: \{task\}\\
- Subtask: \{subtask\}\\
- Error type: \{error type\}\\
- Error stage: \{error stage\}\\
- Error detail: \{error detail\}\\
- Correction suggestion: \{error correction\}\\
- Perturbation ([x, y, z]): \{error low level\}\\
The perturbation is the difference between the actual position of the end-effector and the desired target position when the error occurs, where the X-axis points in front of the manipulator, the Y-axis points to the left, and the Z-axis points up. Namely, if the X-axis is positive, the end-effector is in front of the desired target position and causes the task to fail.\\

According to the video and the information, you need to answer the following questions:\\
1. Explain why the task is failed in detail.\\
2. Give detailed High-level correction instructions to help the robot arm to recover from the failure. The high-level correction should describe what subtask the robot arm should perform to recover from the failure.\\
3. Give detailed Low-level correction instructions to help the robot arm to recover from the failure. The low-level correction should describe which direction and how much the robot arm should move to recover from the failure.\\

Please note that specific numerical values should not be given to describe the extent of the low-level correction.
An example of the low-level correction is: "Move the robot arm backward then move the robot arm to the left to align with the target object".\\
Please note that specific numerical values should not be given in the explanation of the failure reason and the high-level correction, you should instead using rich language to describe the failure reason and the high-level correction.\\

Your answer should be in the following JSON format:\\
\{\\
    "reason": <reason>,\\
    "high level correction": <high level correction>,\\
    "low level correction": <low level correction>\\
\}

\end{AIbox}

\section{Human Data Annotation Details}

This section provides detailed descriptions of the human verification protocol, annotation schema, statistical agreement computation, and dataset filtering results.

\subsection{Annotation Protocol}

All trajectories validated through simulation and LLM-based auto-annotation are subjected to human quality assessment. The dataset is first shuffled and then partitioned for annotation. For consistency evaluation, 10\% of the samples are randomly selected and assigned to two annotators independently. The remaining 90\% of samples are assigned to a single annotator. Annotators are randomly drawn from a shared annotator pool to avoid systematic pairing bias.

Each annotator reviews the simulated trajectory together with its corresponding LLM-generated annotation and evaluates whether the annotation correctly reflects the failure behavior observed in simulation.

\subsection{Annotation Schema}

For each sample, annotators provide:

\paragraph{Quality Rating (0--3, ordinal).}
\begin{itemize}
    \item 0: Completely inconsistent with the trajectory.
    \item 1: Largely inconsistent.
    \item 2: Mostly consistent.
    \item 3: Fully consistent.
\end{itemize}

\paragraph{Confidence Score (1--3).}
\begin{itemize}
    \item 1: Low confidence.
    \item 2: Moderate confidence.
    \item 3: High confidence.
\end{itemize}

Quality ratings are used for consistency analysis and dataset filtering. Samples receiving ratings of 0 or 1 are considered invalid and removed from the final dataset. Confidence scores are used for statistical reporting but do not directly affect filtering decisions.

\subsection{Agreement Metric: Krippendorff's $\alpha$}

We evaluate global consistency using Krippendorff's $\alpha$ with an ordinal distance metric. The coefficient is defined as
\begin{equation}
\alpha = 1 - \frac{D_o}{D_e},
\end{equation}
where $D_o$ and $D_e$ denote observed and expected disagreement, respectively. 
For ordinal ratings, the distance function is
\begin{equation}
\delta(i,j) = (i - j)^2.
\end{equation}

This formulation supports sparse multi-rater annotations with missing entries. We obtain $\alpha = 0.86$, indicating high inter-annotator reliability.

\subsection{Filtering Results}

After human verification, 8,559 trajectories are retained and 3,809 are removed based on quality ratings. The average annotator confidence is 2.6, suggesting generally high subjective certainty in the assigned quality ratings.

\section{Evaluation Details}
\label{evaluation_details}
\textbf{Construct Multiple-Choice Question Options.}
For the evaluation of three distinct question types—\textit{failure Detection}, \textit{failure Identification}, and \textit{failure locating}, we adopt a multiple-choice question format. The construction of answer options for each task is as follows:

\begin{itemize}

\item \textit{Failure detection}: The model selects from a binary choice set: \textbf{<Yes/No>}.

\item \textit{Failure identification}: The model chooses from a predefined set of six failure types: \textbf{['Orientation deviation.', 'Step omission.', 'Wrong target object.', 'Timing error.', 'Grasping error.', 'Position deviation.']}.

\item \textit{Failure locating}: Four sub-stagess are randomly sampled from all the sub-stages in the RoboFAC dataset and combined with the correct sub-stage corresponding to the current sample. These five options are then shuffled to form the final choice set.

\end{itemize}

\textbf{Evaluate by LLM.}
For the remaining five question types—\textit{task identification}, \textit{task planning}, \textit{failure explanation}, \textit{high-level correction}, and \textit{low-level correction}—we evaluate model responses using GPT-4 as a scoring agent. The evaluation is conducted across three dimensions, each rated on a 1–5 scale:
\begin{itemize}
\item Correctness: Factual accuracy and consistency with the reference answer.

\item Relevance: The degree to which the model’s response addresses the given question.

\item Completeness: Whether the response sufficiently covers all key aspects of the reference answer.
\end{itemize}

To ensure fairness and consistency in the scoring results, we configure GPT-4 with a temperature of $0.2$ and a Top-P value of $1.0$. We prompt GPT-4 with the question, the reference answer, and the response generated by the testing model, asking it to assign scores based on the criteria above. The exact prompt used is as follows:

\begin{AIbox}[Prompt for LLM scoring]

You are an expert evaluator. Assess the quality of a model's response to the user's query.\\

Question: \{question\}\\

Reference answer: \{ref\}\\

Model's response: \{pred\}\\

Evaluate the model's response on the following criteria:\\
- correctness: factual accuracy and consistency with the reference answer.\\
- relevance: how well the model's response addresses the question.\\
- completeness: whether all key aspects of the reference answer are covered.\\

For each criterion, provide a score from 0 to 5 and a **brief** explanation, the score should be an integer.
The score you give needs to be strict and demanding.\\

Output ONLY the JSON object in the following format:\\
\{\\
"criteria": \{\\
    "correctness": \{"score": <0-5>, "explanation": <brief explanation>\},\\
    "relevance": \{"score": <0-5>, "explanation": <brief explanation>\},\\
    "completeness": \{"score": <0-5>, "explanation": <brief explanation>\},\\
\}\\
\}

\end{AIbox}

\section{Supplementary Evaluation Results}

We evaluate six models: Qwen-2.5-VL-3B, Qwen-2.5-VL-7B, two proprietary systems (Gemini-2.0, GPT-4o), and our proposed RoboFAC-3B and RoboFAC-7B. This section details their results on the RoboFAC benchmark.

Table~\ref{tab:sim_ques} summarizes the results on simulation evaluation, while Table~\ref{tab:real_ques} provides the results on real-world evaluation.

\begin{table*}[htbp]
\centering
\caption{Model Performance on different question dimensions for simulation dataset.}
\label{tab:sim_ques}
    \adjustbox{max width=\textwidth}{
    \begin{tabular}{lcccccccc}
        \toprule
        \textbf{Model} & \makecell{\textbf{Task} \\ \textbf{identification}} & \makecell{\textbf{Task} \\ \textbf{planning}} & \makecell{\textbf{Failure} \\ \textbf{explanation}} & \makecell{\textbf{High-level} \\ \textbf{correction}} & \makecell{\textbf{Low-level} \\ \textbf{correction}} & \makecell{\textbf{Failure} \\ \textbf{detaction}} & \makecell{\textbf{Failure} \\ \textbf{identification}} & \makecell{\textbf{Failure} \\ \textbf{locating}} \\
        \midrule
        Qwen2.5-VL-3B & 22.619 & 25.530 & 25.714 & 41.241 & 27.157 & 36.839 & 04.114 & 53.179 \\
        Qwen2.5-VL-7B & 21.746 & 18.728 & 17.628 & 20.075 & 16.980 & 50.463 & 26.103 & 22.513 \\
        Gemini-2.0     & 48.038 & 43.002 & 62.945 & 56.136 & 41.824 & 45.966 & 27.076 & 78.459 \\
        GPT-4o         & 39.021 & 45.475 & 42.937 & 57.851 & 46.118 & 65.212 & 21.074 & 70.830 \\
        RoboFAC-3B     & 99.423 & 64.109 & 99.881 & 59.820 & 65.853 & 89.153 & 66.343 & 96.710 \\
        RoboFAC-7B     & 99.907 & 66.213 & 99.784 & 65.979 & 67.245 & 91.270 & 63.800 & 96.933 \\
        \bottomrule
    \end{tabular}
    }
\end{table*}

\begin{table*}[htbp]
    \centering
    \caption{Model performance on different question dimensions for real-world dataset.}
    \label{tab:real_ques}
        \adjustbox{max width=\textwidth}{
        \begin{tabular}{lcccccccc}
            \toprule
            \textbf{Model} & \makecell{\textbf{Task} \\ \textbf{identification}} & \makecell{\textbf{Task} \\ \textbf{planning}} & \makecell{\textbf{Failure} \\ \textbf{explanation}} & \makecell{\textbf{High-level} \\ \textbf{correction}} & \makecell{\textbf{Low-level} \\ \textbf{correction}} & \makecell{\textbf{Failure} \\ \textbf{detaction}} & \makecell{\textbf{Failure} \\ \textbf{identification}} & \makecell{\textbf{Failure} \\ \textbf{locating}} \\
            \midrule
            Qwen2.5-VL-3B & 32.796 & 26.872 & 18.313 & 23.292 & 21.431 & 03.405 & 02.917 & 05.625 \\
            Qwen2.5-VL-7B & 39.291 & 35.581 & 34.201 & 44.667 & 24.242 & 83.389 & 36.042 & 80.938 \\
            Gemini-2.0     & 60.748 & 77.010 & 18.451 & 24.653 & 24.731 & 59.718 & 12.604 & 15.729 \\
            GPT-4o         & 71.013 & 65.825 & 55.681 & 57.819 & 51.313 & 97.176 & 46.042 & 53.958 \\
            RoboFAC-3B     & 60.731 & 67.813 & 49.750 & 54.868 & 61.970 & 80.150 & 42.708 & 81.979 \\
            RoboFAC-7B     & 69.734 & 76.357 & 56.090 & 59.667 & 63.855 & 80.648 & 57.813 & 71.250 \\
            \bottomrule
        \end{tabular}
        }
\end{table*}

\section{Additional Examples of Failure Analysis}
\begin{figure*}[htbp]
    \centering
    \includegraphics[width=\linewidth]{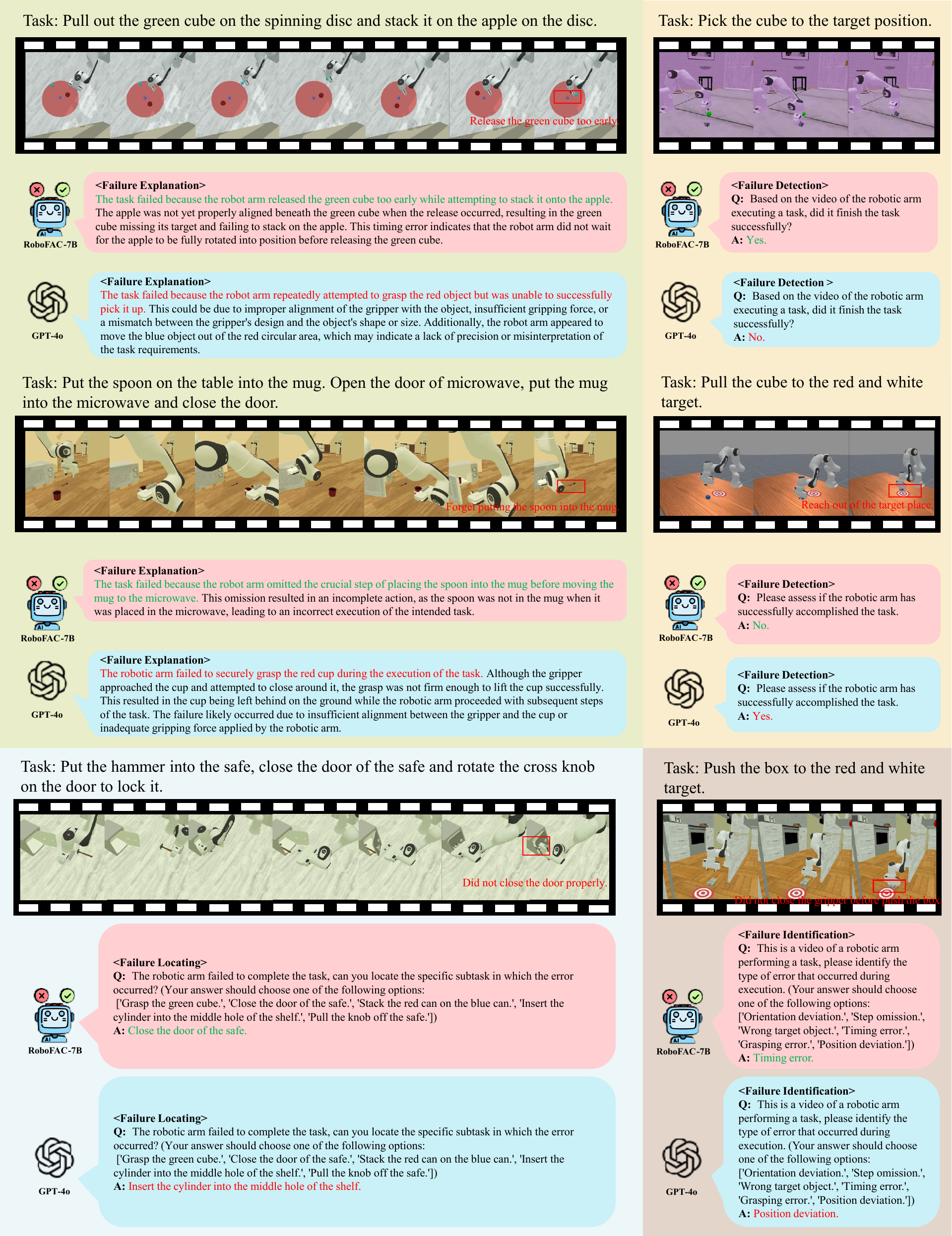}
    \caption{Examples of failure analysis, including failure explanation, detection, locating, and identification. Different background colors are used to indicate different types of questions.}
    \label{fig:DEMO2}
\end{figure*}

\begin{figure*}[htbp]
    \centering
    \includegraphics[width=\linewidth]{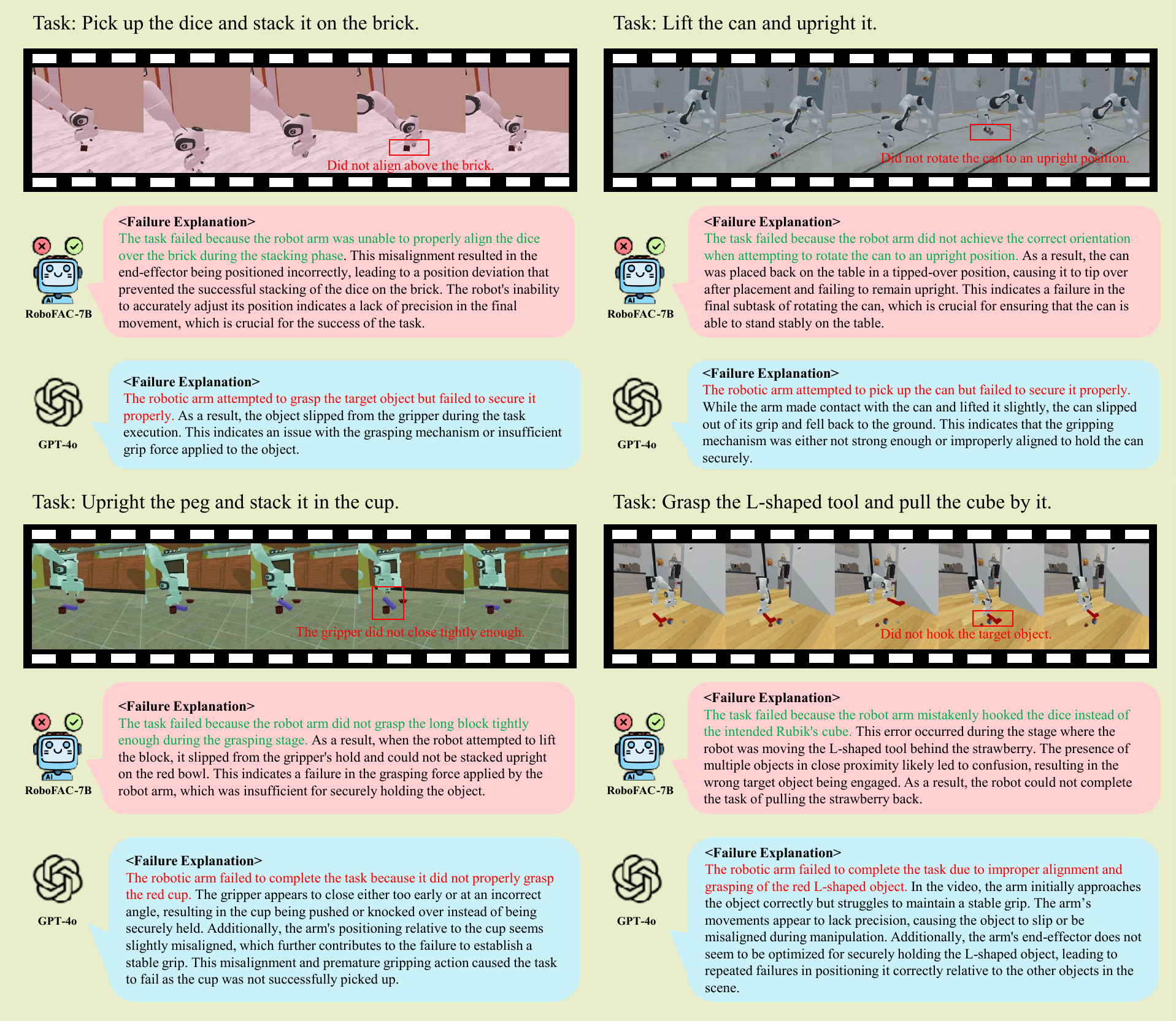}
    \caption{Qualitative comparison of failure explanations generated by RoboFAC-7B and GPT-4o across different tasks.}
    \label{fig:DEMO1}
\end{figure*}

Figure~\ref{fig:DEMO2} further illustrates the multi-dimensional diagnostic capability of RoboFAC-7B. In addition to failure explanation, the model is evaluated on failure detection, locating the specific step where the failure occurred, and identifying the type of error. In all cases, RoboFAC-7B provides correct answers, while GPT-4o fails to correctly diagnose the failures, highlighting the robustness of our model in understanding and analyzing real-world robotic errors.

Figure~\ref{fig:DEMO1} presents several examples comparing the failure explanations generated by RoboFAC-7B and GPT-4o. RoboFAC-7B consistently produces more accurate and concise explanations, correctly identifying the critical steps that caused the failures.

\section{Demos of Failure Correction in Real-world tasks}
\begin{figure*}[htbp]
    \centering
    \includegraphics[width=\linewidth]{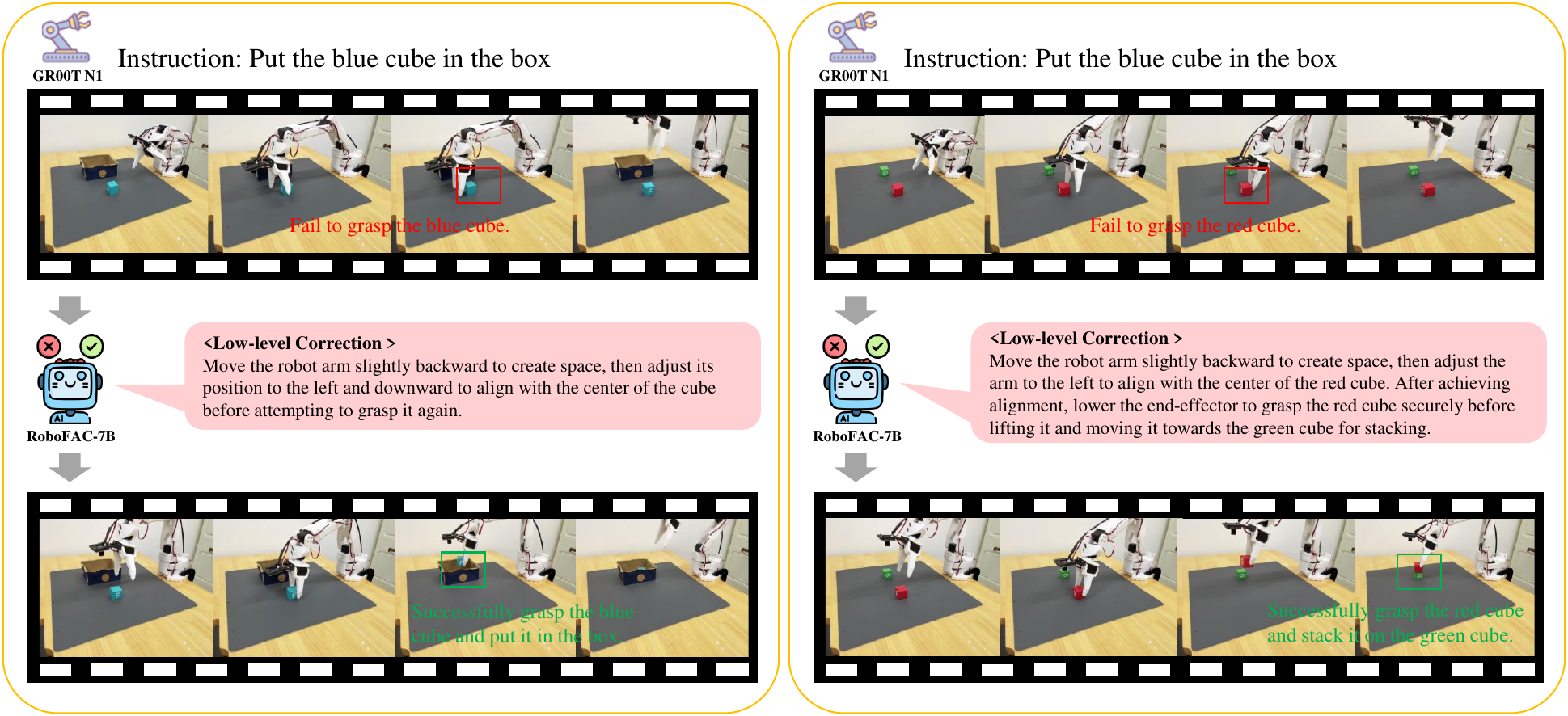}
    \caption{Demo of failure correction in real-world tasks.}
    \label{fig:real_world_task}
\end{figure*}

Figure~\ref{fig:real_world_task} presents two real-world examples demonstrating the effectiveness of RoboFAC-7B in correcting manipulation failures. In both cases, the robot (\textsc{GR00T N1}) initially fails to grasp the target object due to inaccurate alignment. Based on the instruction and visual observations, RoboFAC-7B generates low-level corrective feedback, which guides the robot to adjust its pose and retry the action. The corrected executions successfully complete the task objectives: placing a blue cube into a box (left) and stacking a red cube onto a green one (right).

\end{document}